\documentclass[11pt,a4paper]{article}
\usepackage[hyperref]{acl2019}
\usepackage{times}
\usepackage{latexsym}
\usepackage{amsmath,amsfonts,amssymb,amsthm,bm}
\usepackage{multirow}
\usepackage{url}
\usepackage{array}
\usepackage{graphicx}
\usepackage{subcaption}
\usepackage{amsthm}
\usepackage{amsfonts}
\usepackage{color}
\usepackage{MnSymbol}
\usepackage{makecell}
\usepackage{arydshln}
\usepackage{amsmath}
\usepackage{algorithm}
\usepackage[noend]{algpseudocode}
\usepackage{algorithmicx}
\usepackage[shortlabels]{enumitem}

\newcommand\blfootnote[1]{%
  \begingroup
  \renewcommand\thefootnote{}\footnote{#1}%
  \addtocounter{footnote}{-1}%
  \endgroup
}

\DeclareMathOperator*{\myotimes}{\raisebox{-1pt}{\scalebox{1.44}{$\bigotimes$}}}

\aclfinalcopy 
\def\aclpaperid{2315} 

\title{Learning Representations from Imperfect Time Series Data\\ via Tensor Rank Regularization}

\author{Paul Pu Liang$^{\clubsuit*}$, Zhun Liu$^{\diamondsuit*}$, Yao-Hung Hubert Tsai$^\clubsuit$\\ {\bf Qibin Zhao$^{\heartsuit}$, Ruslan Salakhutdinov$^\clubsuit$, Louis-Philippe Morency$^\diamondsuit$}\\
$^\clubsuit$Machine Learning Department, Carnegie Mellon University, USA\\
$^\diamondsuit$Language Technologies Institute, Carnegie Mellon University, USA\\
$^\heartsuit$Tensor Learning Unit, RIKEN Center for Artificial Intelligence Project, Japan\\
{\tt \{pliang,zhunl,yaohungt,rsalakhu,morency\}@cs.cmu.edu}\\ {\tt qibin.zhao@riken.jp}
}

\date{}

\begin{document}
\maketitle

\begin{abstract}
There has been an increased interest in multimodal language processing including multimodal dialog, question answering, sentiment analysis, and speech recognition. However, naturally occurring multimodal data is often \textit{imperfect} as a result of imperfect modalities, missing entries or noise corruption. To address these concerns, we present a regularization method based on \textit{tensor rank minimization}. Our method is based on the observation that high-dimensional multimodal time series data often exhibit correlations across time and modalities which leads to low-rank tensor representations. However, the presence of noise or incomplete values breaks these correlations and results in tensor representations of higher rank. We design a model to learn such tensor representations and effectively regularize their rank. Experiments on multimodal language data show that our model achieves good results across various levels of imperfection.\blfootnote{$^*$first two authors contributed equally}
\end{abstract}

\vspace{-1mm}
\section{Introduction}
\vspace{-1mm}

Analyzing multimodal language sequences spans various fields including multimodal dialog~\citep{visdial,Rudnicky2005}, question answering~\citep{VQA,DBLP:journals/corr/TapaswiZSTUF15,embodiedqa}, sentiment analysis~\citep{morency2011towards}, and speech recognition~\cite{pala2019}. Generally, these multimodal sequences contain heterogeneous sources of information across the language, visual and acoustic modalities. For example, when instructing robots, these machines have to comprehend our verbal instructions and interpret our nonverbal behaviors while grounding these inputs in their visual sensors~\citep{DBLP:journals/corr/abs-1710-09483,doi:10.1177/0278364904049250}. Likewise, comprehending human intentions requires integrating human language, speech, facial behaviors, and body postures~\citep{Mihalcea:2012:MSA:2392963.2392965,Rossiter2011MultimodalIR}. However, as much as more modalities are required for improved performance, we now face a challenge of \textit{imperfect} data where data might be 1) incomplete due to mismatched modalities or sensor failure, or 2) corrupted with random or structured noise. As a result, an important research question involves learning robust representations from imperfect multimodal data.

\begin{figure}[t!]
    \centering
    \includegraphics[width=\linewidth]{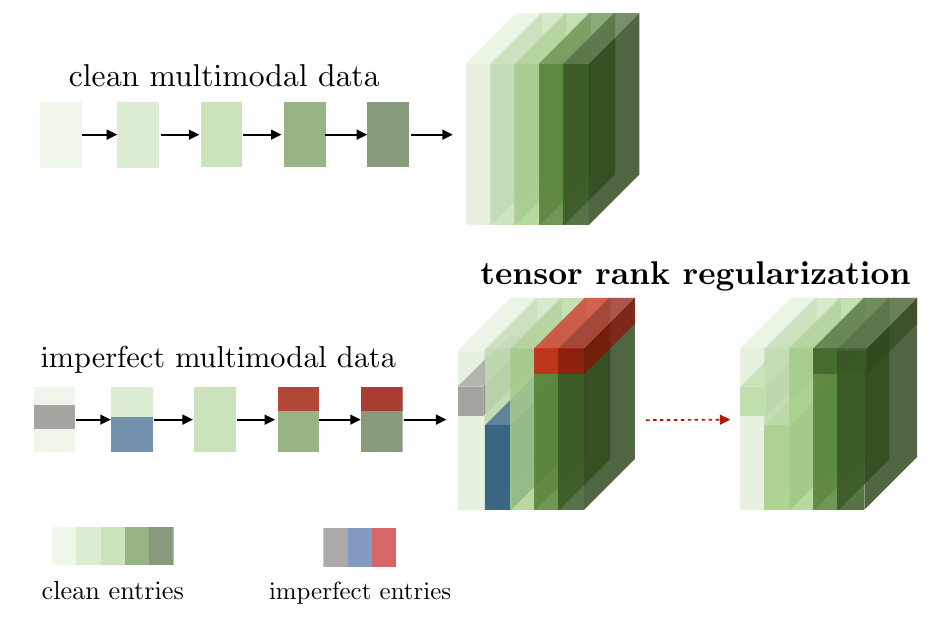}
    \caption{Clean multimodal time series data (in shades of green) exhibits correlations across time and across modalities, leading to redundancy in low rank tensor representations. On the other hand, the presence of imperfect entries (in gray, blue, and red) breaks these correlations and leads to higher rank tensors. In these scenarios, we use \textit{tensor rank regularization} to learn tensors that more accurately represent the true correlations and latent structures in multimodal data.\vspace{-4mm}}
    \label{noise}
\end{figure}

Recent research in both unimodal and multimodal learning has investigated the use of tensors for representation learning~\citep{Anandkumar:2014:TDL:2627435.2697055}. Given representations $\mathbf{h}_1, ..., \mathbf{h}_M$ from $M$ modalities, the order-$M$ outer product tensor $\mathcal{T} = \mathbf{h}_1 \otimes \mathbf{h}_2 \otimes ... \otimes \mathbf{h}_M$ is a natural representation for all possible interactions between the modality dimensions~\citep{lowrank}. In this paper, we propose a model called the Temporal Tensor Fusion Network (T2FN) that builds tensor representations from multimodal time series data. T2FN learns a tensor representation that captures multimodal interactions across time. A key observation is that clean data exhibits tensors that are \textit{low-rank} since high-dimensional real-world data is often generated from lower dimensional latent structures~\citep{Lakshmanan2015ExtractingLL}. Furthermore, clean multimodal time series data exhibits correlations across time and across modalities~\citep{Yang_2017_CVPR,Hidaka:2010:AMT:1891903.1891968}. This leads to redundancy in these overparametrized tensors which explains their low rank (Figure~\ref{noise}). On the other hand, the presence of noise or incomplete values breaks these natural correlations and leads to higher rank tensor representations. As a result, we can use \textit{tensor rank minimization} to learn tensors that more accurately represent the true correlations and latent structures in multimodal data, thereby alleviating imperfection in the input. With these insights, we show how to integrate tensor rank minimization as a simple regularizer for training in the presence of imperfect data. As compared to previous work on imperfect data~\citep{NIPS2014_5279,JMLR:v15:srivastava14b,pham2018found}, our model does not need to know which of the entries or modalities are imperfect beforehand. Our model combines the strength of temporal non-linear transformations of multimodal data with a simple regularization technique on tensor structures. We perform experiments on multimodal video data consisting of humans expressing their opinions using a combination of language and nonverbal behaviors. Our results back up our intuitions that imperfect data increases tensor rank. Finally, we show that our model achieves good results across various levels of imperfection.

\vspace{-1mm}
\section{Related Work}
\vspace{-1mm}

\textbf{Tensor Methods}: Tensor representations have been used for learning discriminative representations in unimodal and multimodal tasks. Tensors are powerful because they can capture important higher order interactions across time, feature dimensions, and multiple modalities~\citep{DBLP:journals/corr/KossaifiLKFA17}. For unimodal tasks, tensors have been used for part-of-speech tagging~\citep{NIPS2014_5323}, dependency parsing~\citep{lei2014low}, word segmentation~\citep{P14-1028}, question answering~\citep{Qiu:2015:CNT:2832415.2832431}, and machine translation~\citep{P15-1004}. For multimodal tasks,~\citet{DBLP:journals/corr/abs-1709-09118} used tensor products between images and text features for image captioning. A similar approach was proposed to learn representations across text, visual, and acoustic features to infer speaker sentiment~\citep{lowrank,tensoremnlp17}. Other applications include multimodal machine translation~\citep{DBLP:journals/corr/DelbrouckD17}, audio-visual speech recognition~\citep{Zhang:2017:TDC:3119899.3063593}, and video semantic analysis~\citep{4907041,GAO2009372}.

\textbf{Imperfect Data}: In order to account for imperfect data, several works have proposed generative approaches for multimodal data~\citep{NIPS2014_5279,JMLR:v15:srivastava14b}. Recently, neural models such as cascaded residual autoencoders~\citep{DBLP:conf/cvpr/Tran0ZJ17}, deep adversarial learning~\citep{Cai:2018:DAL:3219819.3219963}, or translation-based learning~\citep{pham2018found} have also been proposed. However, these methods often require knowing which of the entries or modalities are imperfect beforehand. While there has been some work on using low-rank tensor representations for imperfect data~\citep{DBLP:journals/corr/abs-1709-00192,lowrankimage,DBLP:journals/corr/ChenHWZMLT17,DBLP:journals/corr/abs-1805-03967,NIPS2018_7793}, our approach is the first to integrate rank minimization with neural networks for multimodal language data, thereby combining the strength of non-linear transformations with the mathematical foundations of tensor structures.

\vspace{-1mm}
\section{Proposed Method}
\vspace{-1mm}

In this section, we present our method for learning representations from imperfect human language across the language, visual, and acoustic modalities. In \S\ref{p0}, we discuss some background on tensor ranks. We outline our method for learning tensor representations via a model called Temporal Tensor Fusion Network (T2FN) in \S\ref{p1}. In \S\ref{p2}, we investigate the relationship between tensor rank and imperfect data. Finally, in \S\ref{p3}, we show how to regularize our model using tensor rank minimization.

We use lowercase letters $x \in \mathbb{R}$ to denote scalars, boldface lowercase letters $\mathbf{x} \in \mathbb{R}^d$ to denote vectors, and boldface capital letters $\mathbf{X} \in \mathbb{R}^{d_1 \times d_2}$ to denote matrices. Tensors, which we denote by calligraphic letters $\mathcal{X}$, are generalizations of matrices to multidimensional arrays. An order-$M$ tensor has $M$ dimensions, $\mathcal{X} \in \mathbb{R}^{d_1 \times ... \times d_M}$. We use $\otimes$ to denote outer product between vectors.

\vspace{-1mm}
\subsection{Background: Tensor Rank}
\label{p0}

The rank of a tensor measures how many vectors are required to reconstruct the tensor. Simple tensors that can be represented as outer products of vectors have lower rank, while complex tensors have higher rank. To be more precise, we define the rank of a tensor using Canonical Polyadic (CP)-decomposition~\citep{Carroll1970}. For an order-$M$ tensor $\mathcal{X} \in \mathbb{R}^{d_1 \times ... \times d_M}$, there exists an exact decomposition into vectors $\mathbf{w}$:
\begin{equation}
\label{tensor_form}
    \mathcal{X} = \sum_{i=1}^{r} \myotimes_{m=1}^M \mathbf{w}^{i}_{m}.
\end{equation}
The minimal $r$ for exact decomposition is called the \textit{rank} of the tensor. The vectors $\{\{\mathbf{w}^{i}_{m}\}_{m=1}^M\}_{i=1}^r$ are called the rank $r$ decomposition factors of $\mathcal{X}$.

\begin{figure}[t!]
    \centering
    \includegraphics[width=1.4\linewidth]{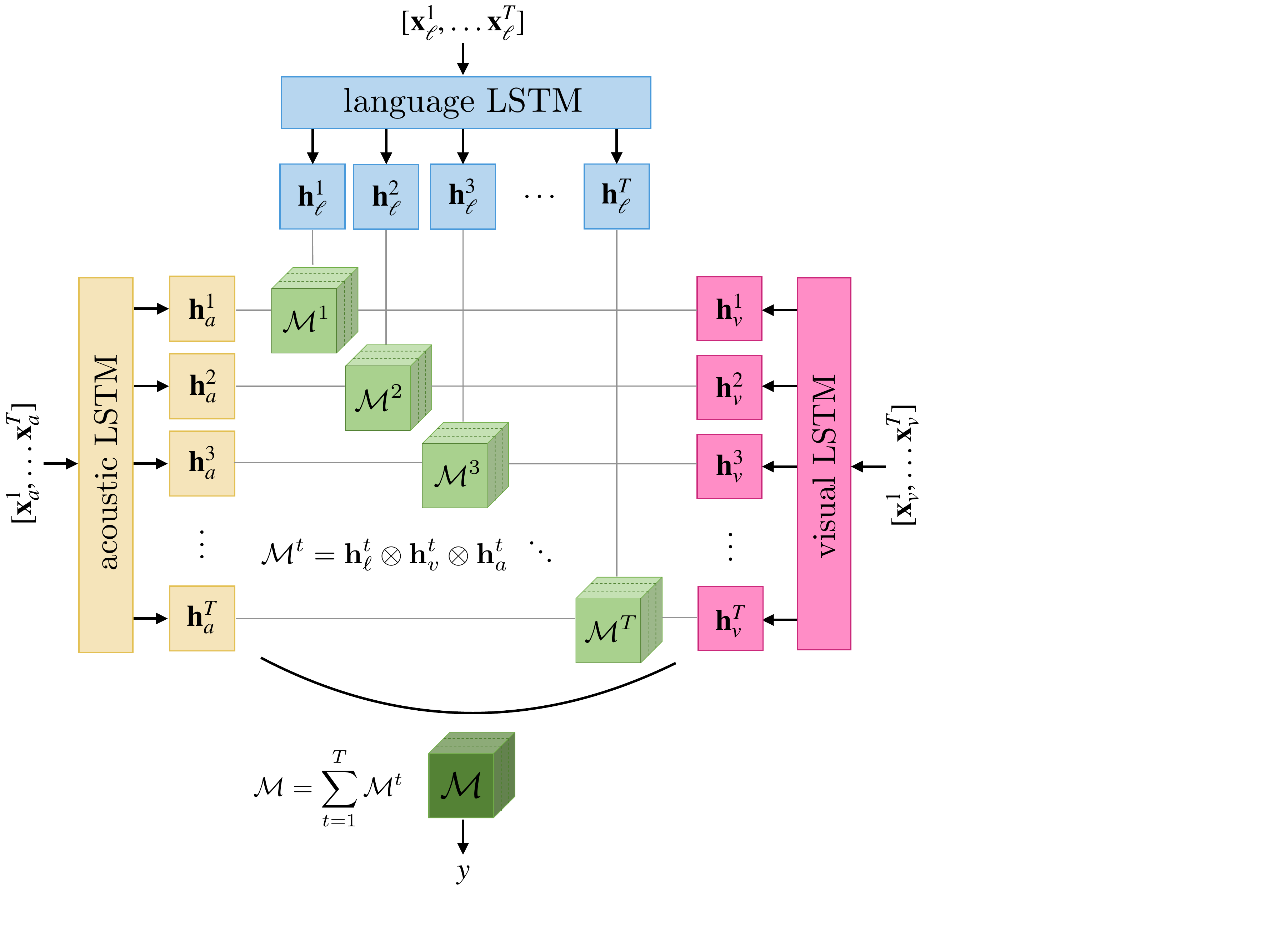}
    \vspace{-10mm}
    \caption{The Temporal Tensor Fusion Network (T2FN) creates a tensor $\mathcal{M}$ from temporal data. The rank of $\mathcal{M}$ increases with imperfection in data so we regularize our model by minimizing an upper bound on the rank of $\mathcal{M}$.\vspace{-4mm}}
    \label{model}
\end{figure}

\vspace{-1mm}
\subsection{Multimodal Tensor Representations}
\label{p1}

Our model for creating tensor representations is called the Temporal Tensor Fusion Network (T2FN), which extends the model in~\citet{tensoremnlp17} to include a temporal component. We show that T2FN increases the capacity of TFN to capture high-rank tensor representations, which itself leads to improved prediction performance. More importantly, our knowledge about tensor rank properties allows us to regularize our model effectively for imperfect data.

We begin with time series data from the language, visual and acoustic modalities, denoted as $[\mathbf{x}_\ell^1, ... , \mathbf{x}_\ell^T]$, $[\mathbf{x}_v^1, ... , \mathbf{x}_v^T]$, and $[\mathbf{x}_a^1, ... , \mathbf{x}_a^T]$ respectively. We first use Long Short-term Memory (LSTM) networks~\citep{hochreiter1997long} to encode the temporal information from each modality, resulting in a sequence of hidden representations $[\mathbf{h}_\ell^1, ... , \mathbf{h}_\ell^T]$, $[\mathbf{h}_v^1, ... , \mathbf{h}_v^T]$, and $[\mathbf{h}_a^1, ... , \mathbf{h}_a^T]$. Similar to prior work which found tensor representations to capture higher-order interactions from multimodal data~\citep{lowrank,tensoremnlp17,fukui2016multimodal}, we form tensors via outer products of the individual representations through time (as shown in Figure~\ref{model}):
\begin{equation}
\label{tensor}
    \mathcal{M} = \sum_{t=1}^{T} \begin{bmatrix} \mathbf{h}_{\ell}^t \\ 1 \end{bmatrix} \otimes \begin{bmatrix} \mathbf{h}_{v}^t \\ 1 \end{bmatrix} \otimes \begin{bmatrix} \mathbf{h}_{a}^t \\ 1 \end{bmatrix}
\end{equation}
where we append a 1 so that unimodal, bimodal, and trimodal interactions are all captured as described in~\citet{tensoremnlp17}. $\mathcal{M}$ is our multimodal representation which can then be used to predict the label $y$ using a fully connected layer. Observe how the construction of $\mathcal{M}$ closely resembles equation~\eqref{tensor_form} as the sum of vector outer products. As compared to TFN which uses a single outer product to obtain a multimodal tensor of rank one, T2FN creates a tensor of high rank (upper bounded by $T$). As a result, the notion of \textit{rank} naturally emerges when we reason about the properties of $\mathcal{M}$.

\vspace{-1mm}
\subsection{How Does Imperfection Affect Rank?}
\label{p2}

We first state several observations about the rank of multimodal representation $\mathcal{M}$: 

\noindent 1) $r_{noisy}$: The rank of $\mathcal{M}$ is maximized when data entries are sampled from i.i.d. noise (e.g. Gaussian distributions). This is because this setting leads to no redundancy at all between the feature dimensions across time steps.

\noindent 2) $r_{clean} < r_{noisy}$: Clean real-world data is often generated from lower dimensional latent structures~\citep{Lakshmanan2015ExtractingLL}. Furthermore, multimodal time series data exhibits correlations across time and across modalities~\citep{Yang_2017_CVPR,Hidaka:2010:AMT:1891903.1891968}. This redundancy leads to low-rank tensor representations.

\noindent 3) $r_{clean} < r_{imperfect} < r_{noisy}$: If the data is imperfect, the presence of noise or incomplete values breaks these natural correlations and leads to higher rank tensor representations.

These intuitions are also backed up by several experimental results which are presented in \S\ref{r1}.

\vspace{-1mm}
\subsection{Tensor Rank Regularization}
\label{p3}

Given our intuitions above, it would then seem natural to augment the discriminative objective function with a term to minimize the rank of $\mathcal{M}$. In practice, the rank of an order-$M$ tensor is computed using the nuclear norm $\lVert \mathcal{X} \rVert_{*}$ which is defined as~\citep{DBLP:journals/corr/FriedlandL14},
\begin{equation}
\small
\label{nuclear}
\lVert \mathcal{X} \rVert_{*} = \inf \left\{ \sum_{i=1}^r | \lambda_i | : \mathcal{X} = \sum_{i=1}^r \lambda_i \left( \myotimes_{m=1}^M \mathbf{w}^{i}_{m} \right), \lVert \mathbf{w}^{i}_{m} \rVert = 1, r \in \mathbb{N} \right\}.
\end{equation}
When $M=2$, this reduces to the matrix nuclear norm (sum of singular values). However, computing the rank of a tensor or its nuclear norm is NP-hard for tensors of order $\ge 3$~\citep{DBLP:journals/corr/FriedlandL14}. Fortunately, there exist efficiently computable upper bounds on the nuclear norm and minimizing these upper bounds would also minimize the nuclear norm $\lVert \mathcal{M} \rVert_{*}$. We choose the upper bound as presented in~\citet{2014arXiv1412.2443H}, which upper bounds the nuclear norm with the tensor Frobenius norm scaled by the tensor dimensions:
\begin{equation}
\label{reg}
\lVert \mathcal{M} \rVert_{*} \le \sqrt{\frac{\prod_{i=1}^{M} d_i }{\max \{ d_1, ..., d_M \}}} \lVert \mathcal{M} \rVert_F,
\end{equation}
where the Frobenius norm $\lVert \mathcal{M} \rVert_F$ is defined as the sum of squared entries in $\mathcal{M}$ which is easily computable and convex. Since $\lVert \mathcal{M} \rVert_F$ is easily computable and convex, including this term adds negligible computational cost to the model. We will use this upper bound as a surrogate for the nuclear norm in our objective function. Our objective function is therefore a weighted combination of the prediction loss and the tensor rank regularizer in equation~\eqref{reg}.

\vspace{-1mm}
\section{Experiments}
\vspace{-1mm}

\begin{figure*}[t!]
\centering
\begin{subfigure}[h]{0.32\textwidth}
\includegraphics[width=\linewidth]{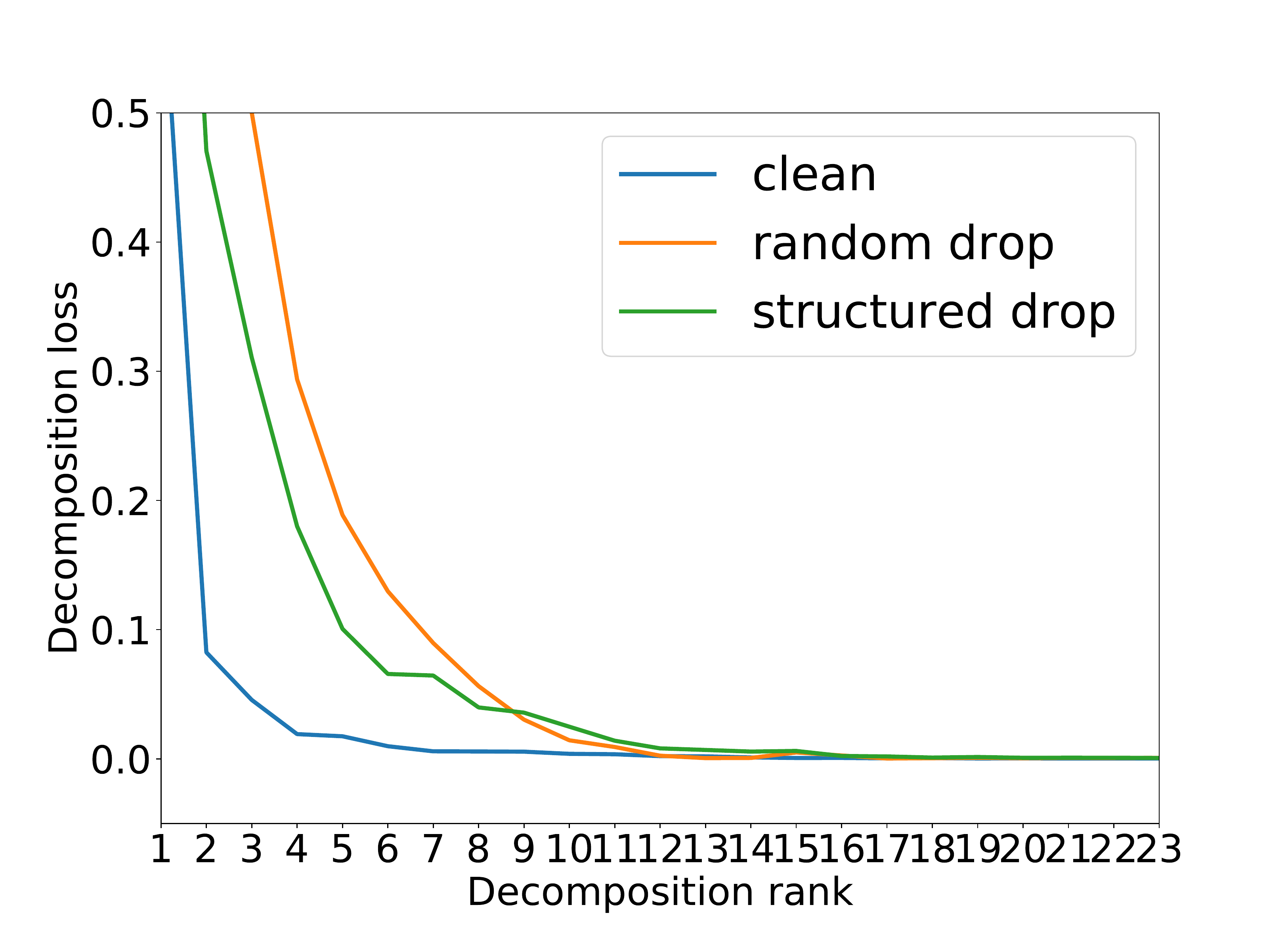}
\caption{CP decomposition error of $\mathcal{M}$ under random and structured dropping of features. Imperfect data leads to an increase in decomposition error and an increase in (approximate) tensor rank.}
\end{subfigure}
\hfill
\begin{subfigure}[h]{0.32\textwidth}
\includegraphics[width=\linewidth]{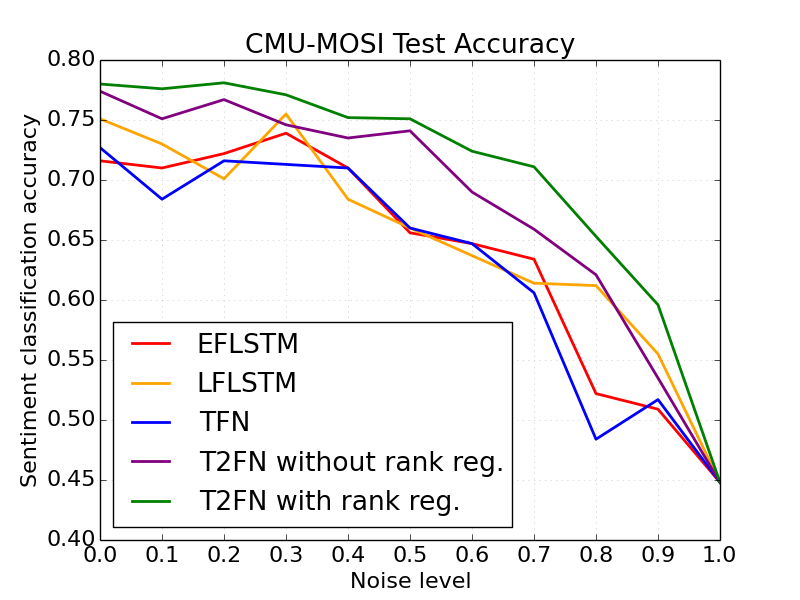}
\caption{Sentiment classification accuracy under random drop (i.e. dropping entries randomly with probability $p \in \texttt{noise\_level}$). T2FN with rank regularization (green) performs well.}
\end{subfigure}
\hfill
\begin{subfigure}[h]{0.32\textwidth}
\includegraphics[width=\linewidth]{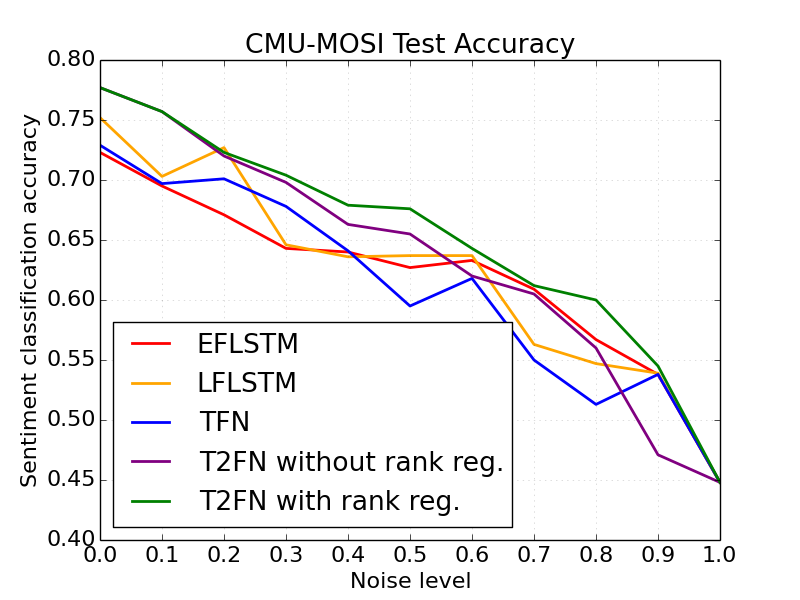}
\caption{Sentiment classification accuracy under structured drop (dropping entire time steps randomly with probability $p \in \texttt{noise\_level}$). T2FN with rank regularization (green) performs well.}
\end{subfigure}
\caption{(a) Effect of imperfect data on tensor rank. (b) and (c): CMU-MOSI test accuracy under imperfect data.\vspace{-4mm}}
\label{pred}
\end{figure*}

Our experiments are designed with two research questions in mind: 1) What is the effect of various levels of imperfect data on tensor rank in T2FN? 2) Does T2FN with rank regularization perform well on prediction with imperfect data? We answer these questions in \S\ref{r1} and \S\ref{r2} respectively.

\vspace{-1mm}
\subsection{Datasets}
\label{sec:data}

We experiment with real video data consisting of humans expressing their opinions using a combination of language and nonverbal behaviors. We use the CMU-MOSI dataset which contains 2199 videos annotated for sentiment in the range $[-3,+3]$~\citep{zadeh2016mosi}. CMU-MOSI and related multimodal language datasets have been studied in the NLP community~\citep{P18-1207,lowrank,liang2018multimodal} from fully supervised settings but not from the perspective of supervised learning with imperfect data. We use 52 segments for training, 10 for validation and 31 for testing. GloVe word embeddings~\citep{pennington2014glove}, Facet~\citep{emotient}, and COVAREP~\citep{degottex2014covarep} features are extracted for the language, visual and acoustic modalities respectively. Forced alignment is performed using P2FA~\citep{P2FA} to align visual and acoustic features to each word, resulting in a multimodal sequence. Our data splits, features, alignment, and preprocessing steps are consistent with prior work on the CMU-MOSI dataset~\citep{lowrank}.

\vspace{-1mm}
\subsection{Rank Analysis}
\label{r1}

We first study the effect of imperfect data on the rank of tensor $\mathcal{M}$. We introduce the following types of noises parametrized by $\texttt{noise\_level} = [0.0,0.1,...,1.0]$. Higher noise levels implies more imperfection: 1) \textbf{clean}: no imperfection, 2) \textbf{random drop}: each entry is dropped independently with probability $p \in \texttt{noise\_level}$, and 3) \textbf{structured drop}: independently for each modality, each time step is chosen with probability $p \in \texttt{noise\_level}$. If a time step is chosen, all feature dimensions at that time step are dropped. For all imperfect settings, features are dropped during both training and testing.

We would like to show how the tensor ranks vary under different imperfection settings. However, as is mentioned above, determining the exact rank of a tensor is an NP-hard problem~\citep{DBLP:journals/corr/FriedlandL14}. In order to analyze the effect of imperfections on tensor rank, we perform CP decomposition (equation~\eqref{tensor_decomp}) on the tensor representations under different rank settings $r$ and measure the reconstruction error $\epsilon$,
\begin{equation}
\label{tensor_decomp}
    \epsilon = \min_{\mathbf{w}^{i}_{m}} \left\lVert \left( \sum_{i=1}^{r} \myotimes_{m=1}^M \mathbf{w}^{i}_{m} \right) - \mathcal{X} \right\rVert_F.
\end{equation}
Given the true rank $r^*$, $\epsilon$ will be high at ranks $r < r^*$, while $\epsilon$ will be approximately zero at ranks $r \ge r^*$ (for example, a rank $3$ tensor would display a large reconstruction error with CP decomposition at rank $1$, but would show almost zero error with CP decomposition at rank $3$). By analyzing the effect of $r$ on $\epsilon$, we are then able to derive a surrogate $\tilde{r}$ to the true rank $r^*$.

Using this approach, we experimented on CMU-MOSI and the results are shown in Figure~\ref{pred}(a). We observe that imperfection leads to an increase in (approximate) tensor rank as measured by reconstruction error (the graph shifts outwards and to the right), supporting our hypothesis that imperfect data increases tensor rank (\S\ref{p2}).

\vspace{-1mm}
\subsection{Prediction Results}
\label{r2}

Our next experiment tests the ability of our model to learn robust representations despite data imperfections. We use the tensor $\mathcal{M}$ for prediction and report binary classification accuracy on CMU-MOSI test set. We compare to several baselines: Early Fusion (EF)-LSTM, Late Fusion (LF)-LSTM, TFN, and T2FN \textit{without} rank regularization. These results are shown in Figure~\ref{pred}(b) for random drop and Figure~\ref{pred}(c) for structured drop. T2FN \textit{with} rank regularization maintains good performance despite imperfections in data. We also observe that our model's improvement is more significant on random drop settings, which results in a higher tensor rank as compared to structured drop settings (from Figure~\ref{pred}(a)). This supports our hypothesis that our model learns robust representations when imperfections that increase tensor rank are introduced. On the other hand, the existing baselines suffer in the presence of imperfect data.

\vspace{-1mm}
\section{Discussion and Future Work}
\vspace{-1mm}

We acknowledge that there are other alternative methods to upper bound the true rank of a tensor~\citep{DBLP:journals/corr/abs-1102-0072,ATKINSON198019,doi:10.1080/03081087.2013.839671}. From a theoretical perspective, there exists a trade-off between the cost of computation and the tightness of approximation. In addition, the tensor rank can (far) exceed the maximum dimension, and a low-rank approximation for tensors may not even exist~\citep{deSilva:2008:TRI:1461964.1461969}. While our tensor rank regularization method seems to work well empirically, there is definitely room for a more thorough theoretical analysis of constructing and regularizing tensor representations for multimodal learning.

\vspace{-1mm}
\section{Conclusion}
\vspace{-1mm}

This paper presented a regularization method based on \textit{tensor rank minimization}. We observe that clean multimodal sequences often exhibit correlations across time and modalities which leads to low-rank tensors, while the presence of imperfect data breaks these correlations and results in tensors of higher rank. We designed a model, the Temporal Tensor Fusion Network, to learn such tensor representations and effectively regularize their rank. Experiments on multimodal language data show that our model achieves good results across various levels of imperfections. We hope to inspire future work on regularizing tensor representations of multimodal data for robust prediction in the presence of imperfect data.

\vspace{-1mm}
\section*{Acknowledgements}
\vspace{-1mm}

PPL, ZL, and LM are partially supported by the National Science Foundation (Award \#1750439 and \#1722822) and Samsung. Any opinions, findings, conclusions or recommendations expressed in this material are those of the author(s) and do not necessarily reflect the views of Samsung and NSF, and no official endorsement should be inferred. YHT and RS are supported in part by DARPA HR00111990016, AFRL FA8750-18-C-0014, NSF IIS1763562, Apple, and Google focused award. QZ is supported by JSPS KAKENHI (Grant No. 17K00326). We also acknowledge NVIDIA's GPU support and the anonymous reviewers for their constructive comments.

\bibliography{acl2019}

\begin{thebibliography}{54}
\expandafter\ifx\csname natexlab\endcsname\relax\def\natexlab#1{#1}\fi

\bibitem[{Alexeev et~al.(2011)Alexeev, Forbes, and
  Tsimerman}]{DBLP:journals/corr/abs-1102-0072}
Boris Alexeev, Michael~A. Forbes, and Jacob Tsimerman. 2011.
\newblock \href {http://arxiv.org/abs/1102.0072} {Tensor rank: Some lower and
  upper bounds}.
\newblock \emph{CoRR}, abs/1102.0072.

\bibitem[{Anandkumar et~al.(2014)Anandkumar, Ge, Hsu, Kakade, and
  Telgarsky}]{Anandkumar:2014:TDL:2627435.2697055}
Animashree Anandkumar, Rong Ge, Daniel Hsu, Sham~M. Kakade, and Matus
  Telgarsky. 2014.
\newblock \href {http://dl.acm.org/citation.cfm?id=2627435.2697055} {Tensor
  decompositions for learning latent variable models}.
\newblock \emph{J. Mach. Learn. Res.}, 15(1):2773--2832.

\bibitem[{Antol et~al.(2015)Antol, Agrawal, Lu, Mitchell, Batra, Zitnick, and
  Parikh}]{VQA}
Stanislaw Antol, Aishwarya Agrawal, Jiasen Lu, Margaret Mitchell, Dhruv Batra,
  C.~Lawrence Zitnick, and Devi Parikh. 2015.
\newblock {VQA}: {V}isual {Q}uestion {A}nswering.
\newblock In \emph{International Conference on Computer Vision (ICCV)}.

\bibitem[{Atkinson and Lloyd(1980)}]{ATKINSON198019}
M.D. Atkinson and S.~Lloyd. 1980.
\newblock \href {https://doi.org/https://doi.org/10.1016/0024-3795(80)90202-5}
  {Bounds on the ranks of some 3-tensors}.
\newblock \emph{Linear Algebra and its Applications}, 31:19 -- 31.

\bibitem[{Ballico(2014)}]{doi:10.1080/03081087.2013.839671}
E.~Ballico. 2014.
\newblock \href {https://doi.org/10.1080/03081087.2013.839671} {An upper bound
  for the real tensor rank and the real symmetric tensor rank in terms of the
  complex ranks}.
\newblock \emph{Linear and Multilinear Algebra}, 62(11):1546--1552.

\bibitem[{Cai et~al.(2018)Cai, Wang, Gao, Shen, and
  Ji}]{Cai:2018:DAL:3219819.3219963}
Lei Cai, Zhengyang Wang, Hongyang Gao, Dinggang Shen, and Shuiwang Ji. 2018.
\newblock \href {https://doi.org/10.1145/3219819.3219963} {Deep adversarial
  learning for multi-modality missing data completion}.
\newblock In \emph{KDD '18}, pages 1158--1166.

\bibitem[{Carroll and Chang(1970)}]{Carroll1970}
J.~Douglas Carroll and Jih-Jie Chang. 1970.
\newblock \href {https://doi.org/10.1007/BF02310791} {Analysis of individual
  differences in multidimensional scaling via an n-way generalization of
  ``eckart-young'' decomposition}.
\newblock \emph{Psychometrika}, 35(3):283--319.

\bibitem[{Chang et~al.(2017)Chang, Yan, Fang, Zhong, and
  Zhang}]{DBLP:journals/corr/abs-1709-00192}
Yi~Chang, Luxin Yan, Houzhang Fang, Sheng Zhong, and Zhijun Zhang. 2017.
\newblock \href {http://arxiv.org/abs/1709.00192} {Weighted low-rank tensor
  recovery for hyperspectral image restoration}.
\newblock \emph{CoRR}, abs/1709.00192.

\bibitem[{Chen et~al.(2017)Chen, Han, Wang, Zhao, Meng, Lin, and
  Tang}]{DBLP:journals/corr/ChenHWZMLT17}
Xiai Chen, Zhi Han, Yao Wang, Qian Zhao, Deyu Meng, Lin Lin, and Yandong Tang.
  2017.
\newblock \href {http://arxiv.org/abs/1705.06755} {A general model for robust
  tensor factorization with unknown noise}.
\newblock \emph{CoRR}, abs/1705.06755.

\bibitem[{Das et~al.(2018)Das, Datta, Gkioxari, Lee, Parikh, and
  Batra}]{embodiedqa}
Abhishek Das, Samyak Datta, Georgia Gkioxari, Stefan Lee, Devi Parikh, and
  Dhruv Batra. 2018.
\newblock {E}mbodied {Q}uestion {A}nswering.
\newblock In \emph{Proceedings of the IEEE Conference on Computer Vision and
  Pattern Recognition (CVPR)}.

\bibitem[{Das et~al.(2017)Das, Kottur, Gupta, Singh, Yadav, Moura, Parikh, and
  Batra}]{visdial}
Abhishek Das, Satwik Kottur, Khushi Gupta, Avi Singh, Deshraj Yadav,
  Jos\'e~M.F. Moura, Devi Parikh, and Dhruv Batra. 2017.
\newblock {V}isual {D}ialog.
\newblock In \emph{Proceedings of the IEEE Conference on Computer Vision and
  Pattern Recognition (CVPR)}.

\bibitem[{Degottex et~al.(2014)Degottex, Kane, Drugman, Raitio, and
  Scherer}]{degottex2014covarep}
Gilles Degottex, John Kane, Thomas Drugman, Tuomo Raitio, and Stefan Scherer.
  2014.
\newblock Covarep - a collaborative voice analysis repository for speech
  technologies.
\newblock In \emph{ICASSP}. IEEE.

\bibitem[{Delbrouck and Dupont(2017)}]{DBLP:journals/corr/DelbrouckD17}
Jean{-}Benoit Delbrouck and St{\'{e}}phane Dupont. 2017.
\newblock \href {http://arxiv.org/abs/1703.08084} {Multimodal compact bilinear
  pooling for multimodal neural machine translation}.
\newblock \emph{CoRR}, abs/1703.08084.

\bibitem[{Fan et~al.(2017)Fan, Chen, Guo, Zhang, and Kuang}]{lowrankimage}
Haiyan Fan, Yunjin Chen, Yulan Guo, Hongyan Zhang, and Gangyao Kuang. 2017.
\newblock \href {https://doi.org/10.1109/JSTARS.2017.2714338} {Hyperspectral
  image restoration using low-rank tensor recovery}.
\newblock \emph{IEEE Journal of Selected Topics in Applied Earth Observations
  and Remote Sensing}, PP:1--16.

\bibitem[{Friedland and Lim(2014)}]{DBLP:journals/corr/FriedlandL14}
Shmuel Friedland and Lek{-}Heng Lim. 2014.
\newblock \href {http://arxiv.org/abs/1410.6072} {Computational complexity of
  tensor nuclear norm}.
\newblock \emph{CoRR}, abs/1410.6072.

\bibitem[{Fukui et~al.(2016)Fukui, Park, Yang, Rohrbach, Darrell, and
  Rohrbach}]{fukui2016multimodal}
Akira Fukui, Dong~Huk Park, Daylen Yang, Anna Rohrbach, Trevor Darrell, and
  Marcus Rohrbach. 2016.
\newblock Multimodal compact bilinear pooling for visual question answering and
  visual grounding.
\newblock \emph{arXiv preprint arXiv:1606.01847}.

\bibitem[{Gao et~al.(2009)Gao, Yang, Tao, and Li}]{GAO2009372}
Xinbo Gao, Yimin Yang, Dacheng Tao, and Xuelong Li. 2009.
\newblock \href {https://doi.org/https://doi.org/10.1016/j.cviu.2008.08.007}
  {Discriminative optical flow tensor for video semantic analysis}.
\newblock \emph{Computer Vision and Image Understanding}, 113(3):372 -- 383.
\newblock Special Issue on Video Analysis.

\bibitem[{Gu et~al.(2018)Gu, Yang, Fu, Chen, Li, and Marsic}]{P18-1207}
Yue Gu, Kangning Yang, Shiyu Fu, Shuhong Chen, Xinyu Li, and Ivan Marsic. 2018.
\newblock \href {http://aclweb.org/anthology/P18-1207} {Multimodal affective
  analysis using hierarchical attention strategy with word-level alignment}.
\newblock In \emph{ACL}.

\bibitem[{Hidaka and Yu(2010)}]{Hidaka:2010:AMT:1891903.1891968}
Shohei Hidaka and Chen Yu. 2010.
\newblock \href {https://doi.org/10.1145/1891903.1891968} {Analyzing multimodal
  time series as dynamical systems}.
\newblock In \emph{International Conference on Multimodal Interfaces and the
  Workshop on Machine Learning for Multimodal Interaction}, ICMI-MLMI '10,
  pages 53:1--53:8, New York, NY, USA. ACM.

\bibitem[{Hochreiter and Schmidhuber(1997)}]{hochreiter1997long}
Sepp Hochreiter and J{\"u}rgen Schmidhuber. 1997.
\newblock Long short-term memory.
\newblock \emph{Neural computation}, 9(8):1735--1780.

\bibitem[{{Hu}(2014)}]{2014arXiv1412.2443H}
Shenglong {Hu}. 2014.
\newblock \href {http://arxiv.org/abs/1412.2443} {{Relations of the Nuclear
  Norms of a Tensor and its Matrix Flattenings}}.
\newblock \emph{arXiv e-prints}, page arXiv:1412.2443.

\bibitem[{Huang et~al.(2017)Huang, Smolensky, He, Deng, and
  Wu}]{DBLP:journals/corr/abs-1709-09118}
Qiuyuan Huang, Paul Smolensky, Xiaodong He, Li~Deng, and Dapeng~Oliver Wu.
  2017.
\newblock \href {http://arxiv.org/abs/1709.09118} {Tensor product generation
  networks}.
\newblock \emph{CoRR}, abs/1709.09118.

\bibitem[{Iba et~al.(2005)Iba, Paredis, and
  Khosla}]{doi:10.1177/0278364904049250}
Soshi Iba, Christiaan J.~J. Paredis, and Pradeep~K. Khosla. 2005.
\newblock \href {https://doi.org/10.1177/0278364904049250} {Interactive
  multimodal robot programming}.
\newblock \emph{The International Journal of Robotics Research}, 24(1):83--104.

\bibitem[{iMotions(2017)}]{emotient}
iMotions. 2017.
\newblock \href {goo.gl/1rh1JN} {Facial expression analysis}.

\bibitem[{Kossaifi et~al.(2017)Kossaifi, Lipton, Khanna, Furlanello, and
  Anandkumar}]{DBLP:journals/corr/KossaifiLKFA17}
Jean Kossaifi, Zachary~C. Lipton, Aran Khanna, Tommaso Furlanello, and Anima
  Anandkumar. 2017.
\newblock \href {http://arxiv.org/abs/1707.08308} {Tensor regression networks}.
\newblock \emph{CoRR}, abs/1707.08308.

\bibitem[{Lakshmanan et~al.(2015)Lakshmanan, Sadtler, Tyler-Kabara, Batista,
  and Yu}]{Lakshmanan2015ExtractingLL}
Karthik Lakshmanan, Patrick~T. Sadtler, Elizabeth~C. Tyler-Kabara, Aaron~P.
  Batista, and Byron~M. Yu. 2015.
\newblock Extracting low-dimensional latent structure from time series in the
  presence of delays.
\newblock \emph{Neural Computation}, 27:1825--1856.

\bibitem[{Lei et~al.(2014)Lei, Xin, Zhang, Barzilay, and Jaakkola}]{lei2014low}
Tao Lei, Yu~Xin, Yuan Zhang, Regina Barzilay, and Tommi Jaakkola. 2014.
\newblock Low-rank tensors for scoring dependency structures.
\newblock In \emph{Proceedings of the 52nd Annual Meeting of the Association
  for Computational Linguistics (Volume 1: Long Papers)}, volume~1, pages
  1381--1391.

\bibitem[{Liang et~al.(2018)Liang, Liu, Zadeh, and
  Morency}]{liang2018multimodal}
Paul~Pu Liang, Ziyin Liu, Amir Zadeh, and Louis-Philippe Morency. 2018.
\newblock Multimodal language analysis with recurrent multistage fusion.
\newblock \emph{EMNLP}.

\bibitem[{Liu et~al.(2018)Liu, Shen, Lakshminarasimhan, Liang, Bagher~Zadeh,
  and Morency}]{lowrank}
Zhun Liu, Ying Shen, Varun~Bharadhwaj Lakshminarasimhan, Paul~Pu Liang, AmirAli
  Bagher~Zadeh, and Louis-Philippe Morency. 2018.
\newblock \href {http://aclweb.org/anthology/P18-1209} {Efficient low-rank
  multimodal fusion with modality-specific factors}.
\newblock In \emph{ACL}.

\bibitem[{Long et~al.(2018)Long, Liu, Chen, and
  Zhu}]{DBLP:journals/corr/abs-1805-03967}
Zhen Long, Yipeng Liu, Longxi Chen, and Ce~Zhu. 2018.
\newblock \href {http://arxiv.org/abs/1805.03967} {Low rank tensor completion
  for multiway visual data}.
\newblock \emph{CoRR}, abs/1805.03967.

\bibitem[{Mihalcea(2012)}]{Mihalcea:2012:MSA:2392963.2392965}
Rada Mihalcea. 2012.
\newblock \href {http://dl.acm.org/citation.cfm?id=2392963.2392965} {Multimodal
  sentiment analysis}.
\newblock In \emph{Proceedings of the 3rd Workshop in Computational Approaches
  to Subjectivity and Sentiment Analysis}, WASSA '12, pages 1--1, Stroudsburg,
  PA, USA. Association for Computational Linguistics.

\bibitem[{Morency et~al.(2011)Morency, Mihalcea, and
  Doshi}]{morency2011towards}
Louis-Philippe Morency, Rada Mihalcea, and Payal Doshi. 2011.
\newblock Towards multimodal sentiment analysis: Harvesting opinions from the
  web.
\newblock In \emph{Proceedings of the 13th international conference on
  multimodal interfaces}, pages 169--176. ACM.

\bibitem[{Nimishakavi et~al.(2018)Nimishakavi, Jawanpuria, and
  Mishra}]{NIPS2018_7793}
Madhav Nimishakavi, Pratik~Kumar Jawanpuria, and Bamdev Mishra. 2018.
\newblock \href
  {http://papers.nips.cc/paper/7793-a-dual-framework-for-low-rank-tensor-completion.pdf}
  {A dual framework for low-rank tensor completion}.
\newblock In S.~Bengio, H.~Wallach, H.~Larochelle, K.~Grauman, N.~Cesa-Bianchi,
  and R.~Garnett, editors, \emph{Advances in Neural Information Processing
  Systems 31}, pages 5484--5495. Curran Associates, Inc.

\bibitem[{Palaskar et~al.(2018)Palaskar, Sanabria, and Metze}]{pala2019}
Shruti Palaskar, Ramon Sanabria, and Florian Metze. 2018.
\newblock End-to-end multimodal speech recognition.
\newblock In \emph{Proceedings of the IEEE International Conference on
  Acoustics, Speech, and Signal Processing (ICASSP)}.

\bibitem[{Pei et~al.(2014)Pei, Ge, and Chang}]{P14-1028}
Wenzhe Pei, Tao Ge, and Baobao Chang. 2014.
\newblock \href {https://doi.org/10.3115/v1/P14-1028} {Max-margin tensor neural
  network for chinese word segmentation}.
\newblock In \emph{Proceedings of the 52nd Annual Meeting of the Association
  for Computational Linguistics (Volume 1: Long Papers)}, pages 293--303.
  Association for Computational Linguistics.

\bibitem[{Pennington et~al.(2014)Pennington, Socher, and
  Manning}]{pennington2014glove}
Jeffrey Pennington, Richard Socher, and Christopher~D Manning. 2014.
\newblock Glove: Global vectors for word representation.
\newblock In \emph{EMNLP}.

\bibitem[{Pham et~al.(2019)Pham, Liang, Manzini, Morency, and
  Poczos}]{pham2018found}
Hai Pham, Paul~Pu Liang, Thomas Manzini, Louis-Philippe Morency, and Barnabas
  Poczos. 2019.
\newblock Found in translation: Learning robust joint representations by cyclic
  translations between modalities.
\newblock \emph{AAAI}.

\bibitem[{Qiu and Huang(2015)}]{Qiu:2015:CNT:2832415.2832431}
Xipeng Qiu and Xuanjing Huang. 2015.
\newblock \href {http://dl.acm.org/citation.cfm?id=2832415.2832431}
  {Convolutional neural tensor network architecture for community-based
  question answering}.
\newblock In \emph{Proceedings of the 24th International Conference on
  Artificial Intelligence}, IJCAI'15, pages 1305--1311. AAAI Press.

\bibitem[{Rossiter(2011)}]{Rossiter2011MultimodalIR}
James Rossiter. 2011.
\newblock Multimodal intent recognition for natural human-robotic interaction.

\bibitem[{Rudnicky(2005)}]{Rudnicky2005}
Alexander~I. Rudnicky. 2005.
\newblock \href {https://doi.org/10.1007/1-4020-3075-4_1} {\emph{Multimodal
  Dialogue Systems}}, pages 3--11. Springer Netherlands, Dordrecht.

\bibitem[{Schmerling et~al.(2017)Schmerling, Leung, Vollprecht, and
  Pavone}]{DBLP:journals/corr/abs-1710-09483}
Edward Schmerling, Karen Leung, Wolf Vollprecht, and Marco Pavone. 2017.
\newblock \href {http://arxiv.org/abs/1710.09483} {Multimodal probabilistic
  model-based planning for human-robot interaction}.
\newblock \emph{CoRR}, abs/1710.09483.

\bibitem[{Setiawan et~al.(2015)Setiawan, Huang, Devlin, Lamar, Zbib, Schwartz,
  and Makhoul}]{P15-1004}
Hendra Setiawan, Zhongqiang Huang, Jacob Devlin, Thomas Lamar, Rabih Zbib,
  Richard Schwartz, and John Makhoul. 2015.
\newblock \href {https://doi.org/10.3115/v1/P15-1004} {Statistical machine
  translation features with multitask tensor networks}.
\newblock In \emph{Proceedings of the 53rd Annual Meeting of the Association
  for Computational Linguistics and the 7th International Joint Conference on
  Natural Language Processing (Volume 1: Long Papers)}, pages 31--41.
  Association for Computational Linguistics.

\bibitem[{de~Silva and Lim(2008)}]{deSilva:2008:TRI:1461964.1461969}
Vin de~Silva and Lek-Heng Lim. 2008.
\newblock \href {https://doi.org/10.1137/06066518X} {Tensor rank and the
  ill-posedness of the best low-rank approximation problem}.
\newblock \emph{SIAM J. Matrix Anal. Appl.}, 30(3):1084--1127.

\bibitem[{Sohn et~al.(2014)Sohn, Shang, and Lee}]{NIPS2014_5279}
Kihyuk Sohn, Wenling Shang, and Honglak Lee. 2014.
\newblock \href
  {http://papers.nips.cc/paper/5279-improved-multimodal-deep-learning-with-variation-of-information.pdf}
  {Improved multimodal deep learning with variation of information}.
\newblock In \emph{NIPS}.

\bibitem[{Srikumar and Manning(2014)}]{NIPS2014_5323}
Vivek Srikumar and Christopher~D Manning. 2014.
\newblock \href
  {http://papers.nips.cc/paper/5323-learning-distributed-representations-for-structured-output-prediction.pdf}
  {Learning distributed representations for structured output prediction}.
\newblock In Z.~Ghahramani, M.~Welling, C.~Cortes, N.~D. Lawrence, and K.~Q.
  Weinberger, editors, \emph{Advances in Neural Information Processing Systems
  27}, pages 3266--3274. Curran Associates, Inc.

\bibitem[{Srivastava and Salakhutdinov(2014)}]{JMLR:v15:srivastava14b}
Nitish Srivastava and Ruslan Salakhutdinov. 2014.
\newblock \href {http://jmlr.org/papers/v15/srivastava14b.html} {Multimodal
  learning with deep boltzmann machines}.
\newblock \emph{JMLR}, 15.

\bibitem[{Tapaswi et~al.(2015)Tapaswi, Zhu, Stiefelhagen, Torralba, Urtasun,
  and Fidler}]{DBLP:journals/corr/TapaswiZSTUF15}
Makarand Tapaswi, Yukun Zhu, Rainer Stiefelhagen, Antonio Torralba, Raquel
  Urtasun, and Sanja Fidler. 2015.
\newblock \href {http://arxiv.org/abs/1512.02902} {Movieqa: Understanding
  stories in movies through question-answering}.
\newblock \emph{CoRR}, abs/1512.02902.

\bibitem[{Tran et~al.(2017)Tran, Liu, Zhou, and Jin}]{DBLP:conf/cvpr/Tran0ZJ17}
Luan Tran, Xiaoming Liu, Jiayu Zhou, and Rong Jin. 2017.
\newblock Missing modalities imputation via cascaded residual autoencoder.
\newblock In \emph{CVPR}.

\bibitem[{{Wu} et~al.(2009){Wu}, {Liu}, and {Zhuang}}]{4907041}
F.~{Wu}, Y.~{Liu}, and Y.~{Zhuang}. 2009.
\newblock \href {https://doi.org/10.1109/TMM.2009.2021724} {Tensor-based
  transductive learning for multimodality video semantic concept detection}.
\newblock \emph{IEEE Transactions on Multimedia}, 11(5):868--878.

\bibitem[{Yang et~al.(2017)Yang, Yumer, Asente, Kraley, Kifer, and
  Lee~Giles}]{Yang_2017_CVPR}
Xiao Yang, Ersin Yumer, Paul Asente, Mike Kraley, Daniel Kifer, and
  C.~Lee~Giles. 2017.
\newblock Learning to extract semantic structure from documents using
  multimodal fully convolutional neural networks.
\newblock In \emph{The IEEE Conference on Computer Vision and Pattern
  Recognition (CVPR)}.

\bibitem[{Yuan and Liberman(2008)}]{P2FA}
Jiahong Yuan and Mark Liberman. 2008.
\newblock Speaker identification on the scotus corpus.
\newblock \emph{Journal of the Acoustical Society of America}.

\bibitem[{Zadeh et~al.(2017)Zadeh, Chen, Poria, Cambria, and
  Morency}]{tensoremnlp17}
Amir Zadeh, Minghai Chen, Soujanya Poria, Erik Cambria, and Louis-Philippe
  Morency. 2017.
\newblock Tensor fusion network for multimodal sentiment analysis.
\newblock In \emph{EMNLP}, pages 1114--1125.

\bibitem[{Zadeh et~al.(2016)Zadeh, Zellers, Pincus, and
  Morency}]{zadeh2016mosi}
Amir Zadeh, Rowan Zellers, Eli Pincus, and Louis-Philippe Morency. 2016.
\newblock Mosi: Multimodal corpus of sentiment intensity and subjectivity
  analysis in online opinion videos.
\newblock \emph{arXiv preprint arXiv:1606.06259}.

\bibitem[{Zhang et~al.(2017)Zhang, Yang, Liu, Chen, and
  Li}]{Zhang:2017:TDC:3119899.3063593}
Qingchen Zhang, Laurence~T. Yang, Xingang Liu, Zhikui Chen, and Peng Li. 2017.
\newblock \href {https://doi.org/10.1145/3063593} {A tucker deep computation
  model for mobile multimedia feature learning}.
\newblock \emph{ACM Trans. Multimedia Comput. Commun. Appl.},
  13(3s):39:1--39:18.

\end{thebibliography}
\bibliographystyle{acl_natbib}

\end{document}